\documentclass[letterpaper, 10 pt, conference]{ieeeconf}  

\usepackage{graphicx}
\usepackage[dvipsnames]{xcolor}
\usepackage[ruled,vlined]{algorithm2e}
\usepackage{algpseudocode}

\usepackage{amssymb}

\usepackage{subcaption}
\usepackage{float}
\IEEEoverridecommandlockouts                              

\usepackage{amsmath} 

\title{\LARGE \bf
Proactive Robot Control for Collaborative Manipulation Using Human Intent}

\author{Zhanibek Rysbek, Siyu Li, Afagh Mehri Shervedani,  Milo\v s \v Zefran
\thanks{This work has been supported by the National Science Foundation grants IIS-1705058, CMMI-1762924, and CCF-2240532.}
\thanks{All the authors are with the Robotics Laboratory,  Department of Electrical and Computer Engineering, the University of Illinois Chicago, Chicago, IL 60607, USA.}%
}

\usepackage[inline]{enumitem}

\begin{document}

\maketitle
\thispagestyle{empty}
\pagestyle{empty}

\begin{abstract}

Collaborative manipulation task often requires negotiation using explicit or implicit communication. An important example is determining where to move when the goal destination is not uniquely specified, and who should lead the motion. This work is motivated by the ability of humans to communicate the desired destination of motion through back-and-forth force exchanges. Inherent to these exchanges is also the ability to dynamically assign a role to each participant, either taking the initiative or deferring to the partner's lead. In this paper, we propose a hierarchical robot control framework that emulates human behavior in communicating a motion destination to a human collaborator and in responding to their actions. At the top level, the controller consists of a set of finite-state machines corresponding to different levels of commitment of the robot to its desired goal configuration. The control architecture is loosely based on the human strategy observed in the human-human experiments, and the key component is a real-time intent recognizer that helps the robot respond to human actions. We describe the details of the control framework, and feature engineering and training process of the intent recognition. The proposed controller was implemented on a UR10e robot (Universal Robots) and evaluated through human studies. The experiments show that the robot correctly recognizes and responds to human input, communicates its intent clearly, and resolves conflict. We report success rates and draw comparisons with human-human experiments to demonstrate the effectiveness of the approach.

\end{abstract}

\section{Introduction}

Robots that physically interact with humans are becoming increasingly available. However, to be useful in practice, such robots must be able to interact with humans in a transparent and predictable way that is similar to how humans interact with other humans. Multi-modality has been an important field of study in Human-Robot Interaction (HRI)~\cite{abbasi_multimodal_2019, monaikul_role_2020, peternel_multimodal_intention_2016, shervedani2023end, gildert_need_2018}. A challenging task within physical HRI (pHRI) is collaborative manipulation, where dyads remain in physical contact throughout and are able to communicate haptically~\cite{groten_haptic_dominance, mortl_role_2012, rysbek_physical_action2021, gildert_need_2018}. 

The specific aspect of the collaborative manipulation task studied in this work is the process through which the participants negotiate where to move, and how to move. The real-world scenario where such a negotiation takes place would be a sudden occurrence of an obstacle that requires the participants to change the direction of motion, with several options available; the participants need to agree on one. Another example would be moving a heavy pot to the table on which several locations for the pot are available; again, the participants need to choose one. While such negotiations often involve several modalities, and in particular verbal communication, none of them except for force interaction is essential. Therefore, in this study, we specifically exclude other communication modalities and investigate how participants use force exchanges to reach a consensus.




\begin{figure*}[t]
    \centering
    \includegraphics[width=\textwidth, trim={0, 5cm, 1cm, 5cm}, clip]{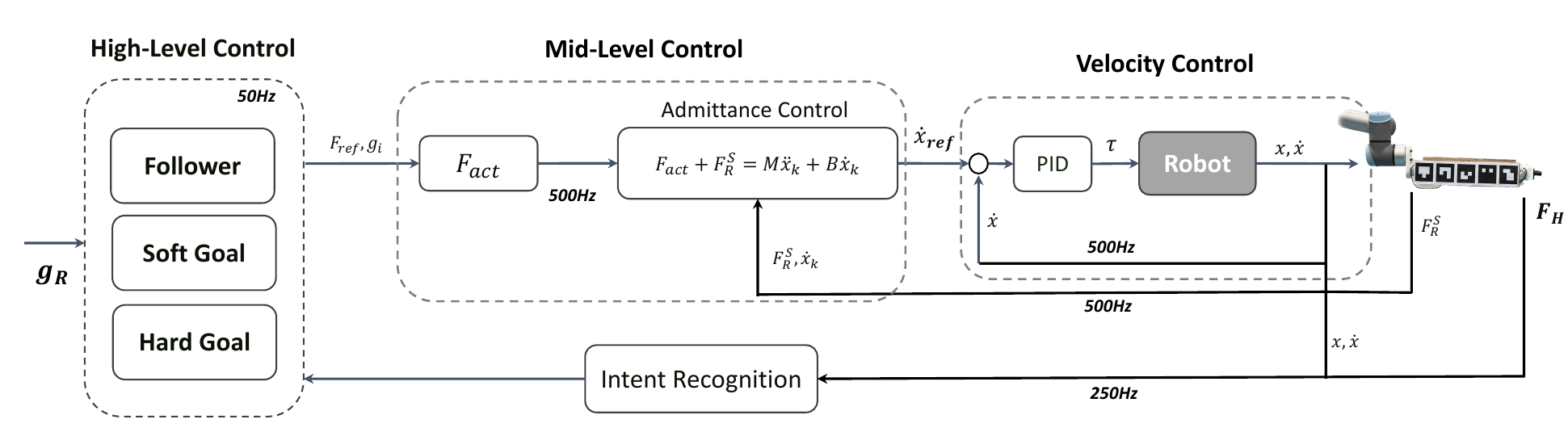}
    \vspace{-5mm}
    \caption{Robot Control Architecture. } 
    \label{fig:control_architecture}
    \vspace{-5mm}
\end{figure*}

Collaborative manipulation has been studied extensively. A substantial body of work uses models of human reaching movements~\cite{MinJerk_FlashHogans1985, kawatoInternalModelsMotor1999a} to control the interaction, often using rule-based intent recognition~\cite{noohi_Model, hamad2021adaptive, duchaine2009safe}. The main objective of such an approach is to assist the human by accelerating or decelerating the manipulated object in order to minimize the effort~\cite{hamad2021adaptive, duchaine2009safe}. A similar approach is to use machine learning techniques to predict \emph{a priori} defined interaction states from haptic signals to integrate them for robot control~\cite{saadiNovelHapticFeature2020, madan_recognition_2015}. Recent work~\cite{alsaadi_resolving_conflicts} proposes a robot controller that detects conflicts from the haptic signals and adapts the robot behavior until the interaction becomes collaborative.

Another popular approach is to use programming by demonstration. Initially targeting point-to-point reaching movements~\cite{Schaal2006_dmp}, the approach has been later successfully adapted to pHRI applications~\cite{ben_amor_interaction_2014, calinon_learning_2009, rozoLearningOptimalControllers2015}. While exhibiting good performance in learning low-level force and velocity characteristics of the task, the interaction with the human is not explicitly controlled and the robot has limited ability to respond to a large variety of human behaviors.


An important aspect of implementing a controller for collaborative manipulation is the low-level control of the point of contact. The predominant approach in the literature for contact-rich tasks is based on impedance/admittance control~\cite{hogan1984impedance}. Specifically for pHRI, \cite{rahman_ikeura_2000, kosugeControlOfRobotHandling1997} designed impedance parameters to match human arm impedance. To adapt to different phases of interaction, impedance gain scheduling is a common practice~\cite{hamad2021adaptive, varol_multiclas_realtime_intent, jacobsonFootContactForces2022, alsaadi_resolving_conflicts}. For human-humanoid collaborative object transportation, several efforts exist~\cite{agravanteCollaborativeHumanHumanoid2014, bussyHumanHumanoidHapticJoint2012} that focus on low-level humanoid control based on human-human experiments for balancing tasks, where equilibrium trajectory position is modulated in the admittance control.

A crucial part of pHRI control is accurate intent recognition. Most of the work is focused on inferring the intended goal based on the kinematic data for tasks that require positional coordination such as grasping, reaching motions and ball catching~\cite{jain_probabilistic_2019,wang_probabilistic_2013,mojtahedi_communication_2017-1, schultzGoalPredictiveRobotic2017}. Getting an accurate human intent from force data remains challenging since the force components are unpredictable due to task specifics. For example, in~\cite{peternel_multimodal_intention_2016}, stiff contact with the environment distorts the underlying intent, hence authors had to use additional sensors. In our case, the interaction forces are influenced by the grasping forces and walking patterns~\cite{rysbek_recognizing_2023}.

 

In most of the existing work on collaborative manipulation, the robot follows the human lead. In contrast, our work focuses on enabling the robot to negotiate where to move, and who will lead the motion. The main contributions of the work are three-fold:
\begin{enumerate*}[label=(\alph*)]
\item A high-level controller that allows the robot to communicate its intent and respond to human actions. The control architecture is motivated by our studies of human-human collaborative manipulation~\cite{rysbek_recognizing_2023}.
\item A real-time intent recognition module for robot control that uses novel force/kinematics features as described in~\cite{rysbek_recognizing_2023}. In contrast, in the related work~\cite{alsaadi_resolving_conflicts, groten_haptic_dominance}, the haptic channel was exclusively used to infer the interaction state rather than the human real-time intent.
\item A human study that validates the framework. The study shows that the framework achieves the stated goals where humans feel that the robot communicates intent, and appropriately responds to their action.
\end{enumerate*}

\section{Approach}


\label{sec:approach}

In collaborative object relocation, there are three important variables to consider: \emph{goal location}, \emph{commitment}, and \emph{what is known} by each agent. The \emph{goal location} is often dependent on the manipulated object. Without loss of generality, we consider generic goal locations $g_i \in SE(2), i=1,\ldots,n$ for a planar task. The \emph{commitment} variable describes how strongly each agent wants to reach the goal. More committed agents tend to escalate the interaction forces while less committed agents may give up their goal and assume a follower role~\cite{rysbek_physical_action2021}. 



In this work, we consider a scenario where each participant is independently assigned a goal (in our case $n=3$) and the level of commitment (\emph{hard}, \emph{soft}, or \emph{none}, meaning that the participant has no goal and needs to follow). A hard goal requires the subject to go to the assigned goal, convincing the partner to comply if necessary. A soft goal should be reached if possible, but if necessary it can be given up and the partner followed instead. Given this task, the robot needs to have three principal controllers:
\begin{enumerate*}
    \item \emph{Follower} controller,
    \item \emph{Soft goal} controller, and
    \item \emph{Hard goal} controller.
\end{enumerate*}
For a detailed description of the task and the human-human experiments see~\cite{rysbek_recognizing_2023}.

\section{Control Architecture}

\subsection{Architecture}


The proposed robot control architecture for the interaction task comprises three distinct layers, as illustrated in Fig.~\ref{fig:control_architecture}. The lower layer employs a Cartesian twist (velocity) controller operating at a frequency of 500Hz. The middle layer implements an admittance controller. The admittance controller provides compliance during the motion. This layer accepts the reference signal from the High-Level Controller (HLC). Finally, the HLC is a set of state machines responsible for imitating human behavior according to roles outlined in Sec.~\ref{sec:approach}. Each state machine is triggered according to a predefined robot goal $g_R$ and responds to human intent feedback (Fig.~\ref{fig:control_architecture}).

 


\subsection{Admittance Control}

Middle-layer controller consists of an admittance control law with inertia and damping terms~\cite{hogan1984impedance,keemink2018admittance}. We used a force-torque sensor mounted between the end-effector and the object to measure the force feedback $F_R^S$. The control equation in the discrete-time domain is:

\begin{equation}
     M \ddot x[k] + B \dot x[k] = F_{act}[k] + F_R^S[k]
\end{equation}
where $M$ and $B$ represent inertia and damping terms, $\dot x[k]$, and $\ddot x[k]$ are the twist and acceleration (in $se(3)$) and $k$ is the time step. Assuming that $\dot x_k = \dot x_{k-1} + \Delta T \ddot x_k$ where $\Delta T$ is the control rate, the inverse dynamics solution for the admittance law is:
\begin{equation}
    \ddot x_k = (M+\Delta TB)^{-1}(F_{act} + F_R^S - B \dot x_{k-1}) \\
\end{equation} When implemented on the robot, a saturation block is added to keep the speed of the robot within safe limits.

\subsection{Robot Action Force}

\begin{figure}
\vspace{2mm}
    \centering
    \includegraphics[width=\columnwidth, trim={2cm, 11cm, 3cm, 11.5cm}, clip]
    {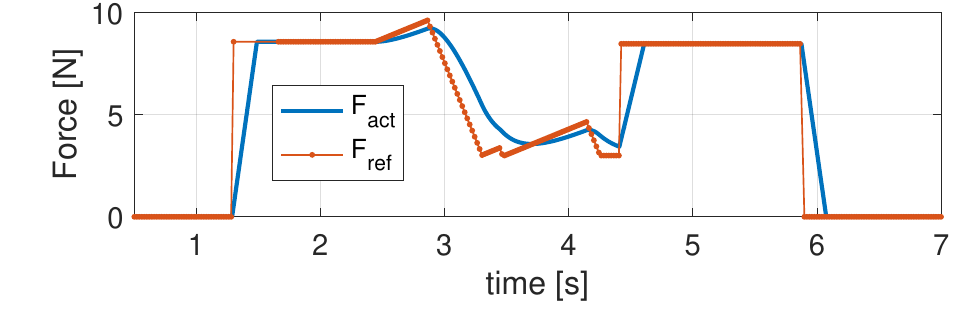}
    \vspace{-5mm}
    \caption{Robot Force Reference Tracking Example.}
    \label{fig:robot_force}
    \vspace{-5mm}
\end{figure}

The robot action force block $F_{act}$ in Fig.~\ref{fig:control_architecture} accepts inputs specifying the desired force magnitude $F_{ref}$ and goal direction $g_i$. In turn, it will interpolate and generate robot action force $F_{act}$ at 500Hz directed towards $g_i$. To avoid sudden force change that might cause aggressive movements, and to be more ergonomic for the user, it is designed to reach the reference point within $T_{transient}$ time period. The value for $T_{transient}$ is chosen to be 0.2 seconds considering the human reaction time~\cite{murchison1934handbook}. Moreover, $F_{act}$ is set to zero when the object reaches any of the goal sites. Robot action force is defined as follows:
\begin{equation}
    F_{act}[k] = F_{act}[k-1] + \frac{F_{ref} - F_{act}[k-1]}{T_{transient}} \Delta T
\end{equation}
Note that, $F_{ref}$ is confined to the safe limits $[F_{min}, F_{max}]$ (see $clip(...)$ in Fig. \ref{fig:state_machines}), hence the magnitude of $F_{act}$ has the same bounds. The dynamic response of $F_{act}$ to $F_{ref}$ can be seen in Fig. \ref{fig:robot_force}.



\subsection{High-Level Reasoning}

\label{sec:hlc}

The high-level control module consists of three state machines corresponding to different types of robot goals as outlined in Sec.~\ref{sec:approach}. Each state machine runs at 50Hz.



\subsubsection{Known Common Goal (KCG)} 

KCG is a basic behavior that is a sub-block of the Follower, Soft and Hard goal controllers. KCG is invoked when the two agents agree on a goal. The controller simply generates a static $F_{ref}$ towards $g_i$ and stops when the target is achieved (Fig.~\ref{fig:state_machines}(a)). Also, it stops at any other goal location in case a user decides to overpower the robot. Thus, it may terminate in two possible states: \emph{Nominal Termination} and \emph{Forced Termination}. It is worth noting that many sophisticated methods exist in literature to provide additional comfort for the user~\cite{hamad2021adaptive, noohi_Model} when the goal location is known to both agents. So, this type of controller is beyond the scope of our work.



 
\subsubsection{Follower}
This module builds on the KCG with an added human intent perception block as shown in Fig.~\ref{fig:state_machines}(b). The human intent perception block accumulates the output of the intent recognizer, and when the timer threshold is met it switches to KCG by selecting the most likely goal. The timer threshold is designed to emulate the open-loop behavior of the human to have enough time to perceive the partner's intent~\cite{kawatoInternalModelsMotor1999a}. Note that the Follower controller does not require $g_R$ as an input. If the human intent is misinterpreted or if it changes after the first round of interaction, the KCG module still has the flexibility to finish with \textit{Forced Termination} state. We note that~\cite{alsaadi_resolving_conflicts} implemented a similar version of this controller but inferred the intended continuous goal location from force directions with a rule-based method.



\subsubsection{Hard Goal}
\begin{figure}
    \centering
    \includegraphics[width=\columnwidth]{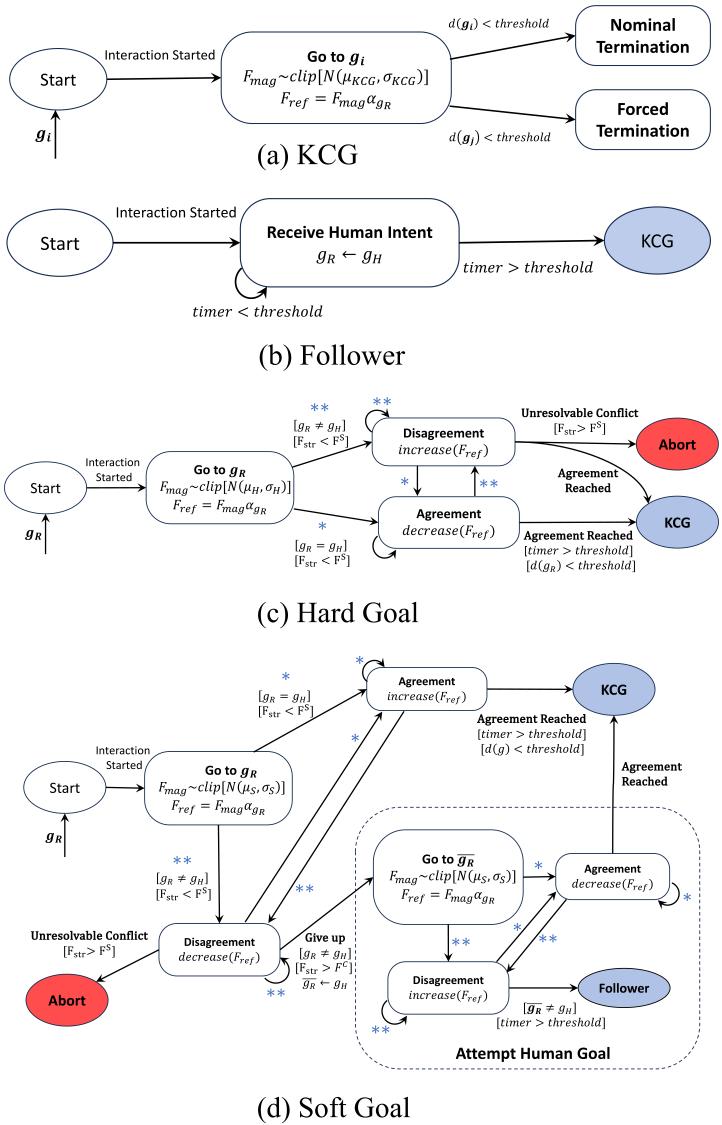}
    \caption{State machines representing the HLC}
    \vspace{-5mm}
    \label{fig:state_machines}
    \vspace{-1mm}
\end{figure}

An important contribution of this paper is the ability of the robot to take initiative and lead the interaction. In the hard goal mode, the robot prioritizes its own goal even though this creates a conflict. Inspired by observations from the human-human experiments~\cite{rysbek_recognizing_2023}, the following logic is implemented as described in Fig.~\ref{fig:state_machines}(c). First, the robot initializes the action force in the direction of the robot's goal. If a conflict is perceived via intent recognition at each control step ($g_R\neq g_H$), then the magnitude of $F_{ref}$ increases, otherwise ($g_R = g_H$) decreases. The rate of increase depends on the current speed towards robot goal $g_R$.

The stretch force $F_{str} = F_H - F_R$ is monitored for safety. If a human applies an excessive amount of force and overpowers the robot, the state machine transitions to the abort state, where the robot will gradually slow down and stop. We note that such situations occurred in human-human experiments. To resolve the conflict in this situation, humans either initiate a dialogue or use other nonverbal cues such as body language, and restart the interaction.

\subsubsection{Soft Goal}

Another contribution of this work is a controller for soft goals. In this mode, the robot can take the initiative or defer to the human. As can be seen in Fig.~\ref{fig:state_machines}(d), the soft goal has an additional subtask \textbf{Attempt Human Goal} (AHG) compared to the hard goal. AHG is triggered if the stretch force $F_{str}$ becomes large (but not large enough for abort). This threshold parameter is denoted as $F^C$. When triggering the AHG, the robot sets its goal to the perceived human intent and invokes the familiar agreement-disagreement cycle. However, if the robot misinterprets the human goal and spends too much time in the disagreement state, the robot switches to the Follower mode, trying to re-interpret human intent. If the robot interprets the human intent correctly then the subtask terminates by transitioning to the KCG mode.


\subsection{Robot Force Sampling}
\label{sec:robot_force_sampling}

Since $F_{ref}$ is the key control variable, it is important to carefully initialize the magnitude of the robot's force. Based on the predefined admittance gain, one can fine-tune the range of the robot force magnitude $F_{mag}=\|F_{ref}\|$ that will act in the spectrum from light to heavy. In this work, we chose to sample $F_{mag}$ from three different regions: weak, medium, and strong. Each level is described by a normal distribution with means ($\mu$) uniformly spread out within $[F_{min}, F_{max}]$ and a fixed standard deviation of $\sigma_S=0.6$N. At the HLC level, $F_{ref}$ is sampled based on the predefined levels above. In addition to that, each controller randomly samples the strength level before initializing $F_{ref}$. For example, the hard goal controller is more likely to sample ($\mu_H$) at a strong level than a soft goal controller ($\mu_S$). The KCG mode mostly samples from the weak level. This variability makes the robot's behavior more human-like.




\subsection{Intent Recognition}

In this work, intent recognition serves both to detect conflict and to determine the human intended goal direction and is the key variable that governs state transitions in HLC. The intent recognition for the human-robot setting was informed by our previous work~\cite{rysbek_recognizing_2023} that studied the human-human interaction. In this work, instead of window intervals, instantaneous values of power signals are used. In order to train the model, several trials of human-robot interaction were collected where the participant moved the object in different goal directions while the robot was in passive admittance control mode. Details of the training procedure are described in Sec.~\ref{sec:intent_recognition_training}.










\section{Implementation}

\subsection{Hardware Setup}

We use a UR10e robot (Universal Robots), and a wooden tray with dimensions 61cm$\times$31cm$\times$10cm and weight $m=2.1$kg. Three goal locations ($n=3$) were chosen to resemble human-human experiments in~\cite{rysbek_recognizing_2023}, with a $40^\circ$ separation angle and 0.5m away from the starting point to accommodate for the reachable workspace of the manipulator. The tray is equipped with two RFT60 force-torque sensors (Robotous), connecting both to an onboard Raspberry Pi and a battery on the back of the tray. The Raspberry Pi transmits the force-torque data to the main computer wirelessly at 200Hz. The force sensor mounted between the robot end-effector and the tray was used as feedback for admittance control. To simplify the implementation, a user-side force sensor was used for intent recognition although the user force could be calculated. To mitigate noise in the feedback system that causes a vibration in admittance control, the real-time Butterworth low-pass filter was implemented with a cutoff frequency of 5Hz. In order to reach the control rate of the robot, force-torque data was up-sampled to 600Hz for smooth admittance control. The Cartesian configuration of the manipulator and its velocity were also streamed at 500Hz. The Robot Operating System (ROS) was used to implement the proposed controller.

Moreover, to synchronize the robot-human joint motion, three beeps are used. The first beep signaled to the human to grasp the object. The second beep was automatically triggered when the sensors detected that the human grasped the tray handle. The third beep was played when the object reached one of the goal locations.






\subsection{Intent Recognition Training}
\label{sec:intent_recognition_training}

Training data for the intent recognition consisted of 18 trials (12 for training, 6 for testing) from two participants. In each trial, the participant attempted to move the tray in a single goal direction and the robot passively followed the human. The distribution of the goals was uniform. The intent recognition module uses only velocity, position, and force (human side) data. Further, goal-projected force, power, velocity, and raw stretch force ($F_{str}$ in object frame) features are used. This is a subset of Feature Set 2 in~\cite{rysbek_recognizing_2023}.
The input to the classifier was a 13-dimensional feature vector sampled at 250Hz. In turn, classifier output was one of the 3 goal locations ($g_1, g_2, g_3$). The idle state (no human action) was determined by a simple threshold on the magnitude of $F_H$. Prior to training, each trial was annotated for an action phase where a participant actively applied force to move the object to a desired location.


\begin{figure}[h]
    \centering
    \includegraphics[width=0.9\columnwidth]{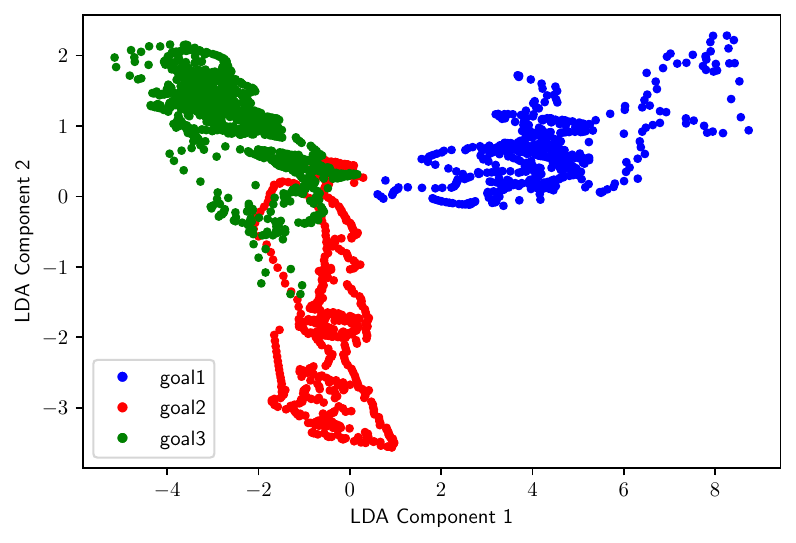} 
    \caption{Visualization of LDA Classification.}
    \label{fig:LDA_vis}
    \vspace{-2mm}
\end{figure}

We chose the Linear Discriminant Analysis (LDA) classification algorithm as our predictor~\cite{hastie2009elements}.
Classification accuracy was 93.47\% for the test set. The separation of classes is shown in Fig.~\ref{fig:LDA_vis} for two principal components. Although the intent recognizer was trained for the passive mode controller, it demonstrated good generalization performance when the robot was actively applying force. Therefore, from the practical point of view, there was no need to generate new data with different robot behaviors. Studying how generalizable the trained model is across different humans is the subject of future work.

\section{Human-Robot Study and Results}

To validate the proposed controller, a human-robot experiment, approved by the IRB, was designed. The task is to collaboratively carry the tray to one of the 3 goal locations. Communication is restricted only to the haptic modality. Before each trial, the human participant was given a goal configuration (location and commitment) as defined in Sec.~\ref{sec:approach}; similarly, a goal configuration was selected for the robot. Both, human and robot agents did not have access to each other's goals. The experimental procedure replicates the human-human study in~\cite{rysbek_recognizing_2023}.

The experiment involved 10 healthy subjects who were recruited from the University of Illinois Chicago campus. Each subject completed 24 trials, totaling 240 trials. The goal assignment for the human-robot dyad was drawn randomly but skewed so more \emph{soft vs. soft} conflicting interactions occurred. Before engaging in the task, the subjects were given time to practice, adapt to the robot, and learn the goal types. This alleviates the learning effect. Following the experiment, the participants were asked to complete a survey for qualitative feedback.





\begin{figure}
    \centering
    \includegraphics[width=\columnwidth, trim={0cm, 0cm, 0cm, 0cm}, clip]{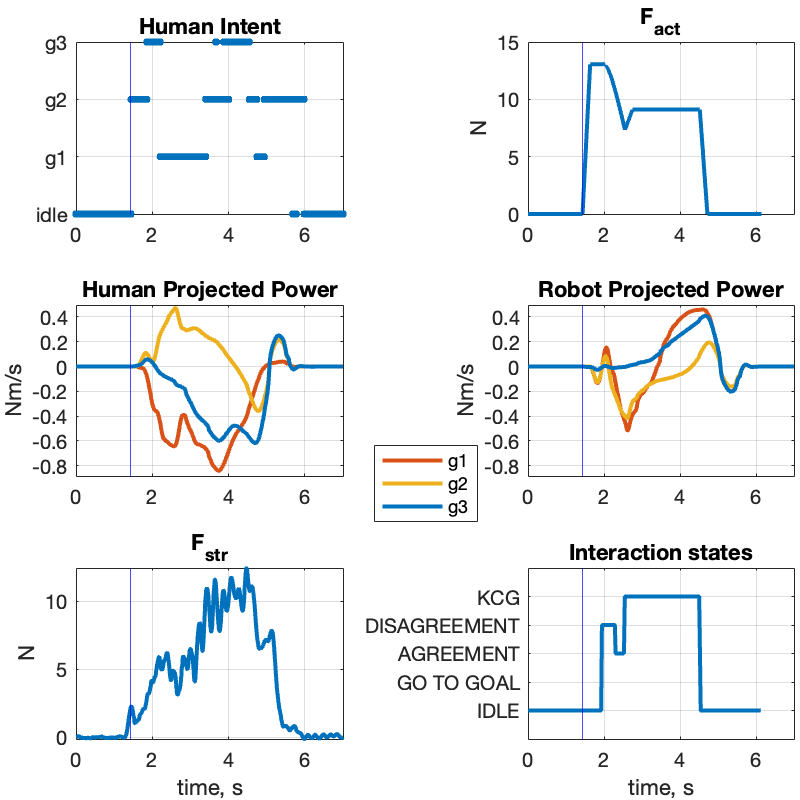}
    \caption{An example of hard-soft case where $g_R=g_1$ and $g_H=g_2$. The blue vertical curve shows the instant of the first beep signal.}
    \label{fig:hard_soft}
    \vspace{-5mm}
\end{figure}



\subsection{Controler Performance}

\subsubsection{Follower Controller}

The number of instances where the robot assumed a follower role in the experiment was 20. In 16 instances, the controller terminated at Normal Termination (Fig.~\ref{fig:state_machines}). This indicates that the intent recognition was correctly determining human input and transitioned to the KCG mode with a correctly recognized common goal. Out of 4 failed cases, 3 were due to humans mistakenly swapping the goal location. One instance failed because the force applied by the human subject was small; the intent recognizer tends to fail when a human is not decisive. During all of the Follower controller instances, the human subject was assigned to different levels of commitment (9 Hard, 11 Soft) and actively applied force on the object. This demonstrates the generalizability and feasibility of the proposed intent-based feedback control scheme. 




\subsubsection{Hard Goal Controller}

In 90 trials, the robot was assigned a hard goal. In 78 instances (87\%), the controller behaved as expected, and the tray was delivered to the robot's intended goal site. In 8 instances (9\%), the controller terminated due to the safety abort. Only in 4 instances (4.5\%) did the robot fail to move the object to the intended goal site. There can be two reasons for such an outcome. An obvious possibility is that the participant failed to perceive the conflict. In fact, in the human-human experiment~\cite{rysbek_recognizing_2023}, human participants failed to perceive the hard goal intent of the partner in several instances. Another reason for a failure can be a small magnitude of $F_{ref}$. In this case, human participants may perceive the robot as having a soft goal, hence the participant decides to apply more effort to bring it to her assigned goal site. As stated in Sec.~\ref{sec:robot_force_sampling}, $F_{ref}$ can vary from trial to trial. 


An example run of a hard goal controller is given in Fig.~\ref{fig:hard_soft}. The robot was assigned to go to $g_1$ and the human to $g_2$. From the human projected power, one can see that the human attempts to go to $g_2$. The intent recognition correctly detects the conflict and the controller increases $F_{act}$. At around 2 seconds, the human partner decides to give up and follow the robot lead; the controller appropriately transitions to the agreement phase. Shortly after, the negotiated goal settles at $g_1$, the controller transitions to the KCG mode and the interaction terminates at $g_1$. Though not used by the controller, one can observe how the robot intent is reflected in the robot's projected power.




\subsubsection{Soft Goal Controller}

Out of 240 trials, the robot had a soft goal in 130 trials and 117 of those were successful (90\%), meaning that: 1) the tray ended up at either the robot's or human's goal if the human had a soft goal too; 2) the tray ended up at the robot's goal if the human had a follower role; 3) the tray ended up at the human's goal if the human had a hard goal. In the remaining 5 instances, the controller stopped due to a safety abort ($\approx 4\%$). In 2 instances ($\approx 2\%$), participants failed to reach the hard goal because they mistakenly swapped the goal location. In the remaining 6 instances ($\approx 5\%$), the dyads misunderstood the intent of the partner in \emph{soft}-\emph{soft} trials. Again, such misunderstandings also occur in human-human interactions.

We also report who prevailed in \emph{soft}-\emph{soft} trials (60 trials). In 34 trials (57\%) the tray ended up at the robot's goal and in 26 trials it ended up at the human's goal. This indicates that the robot controller was able to both take the initiative and defer to the human partner.

\begin{figure}
    \centering
    \includegraphics[width=\columnwidth, trim={0cm, 0cm, 0cm, 0cm}, clip]{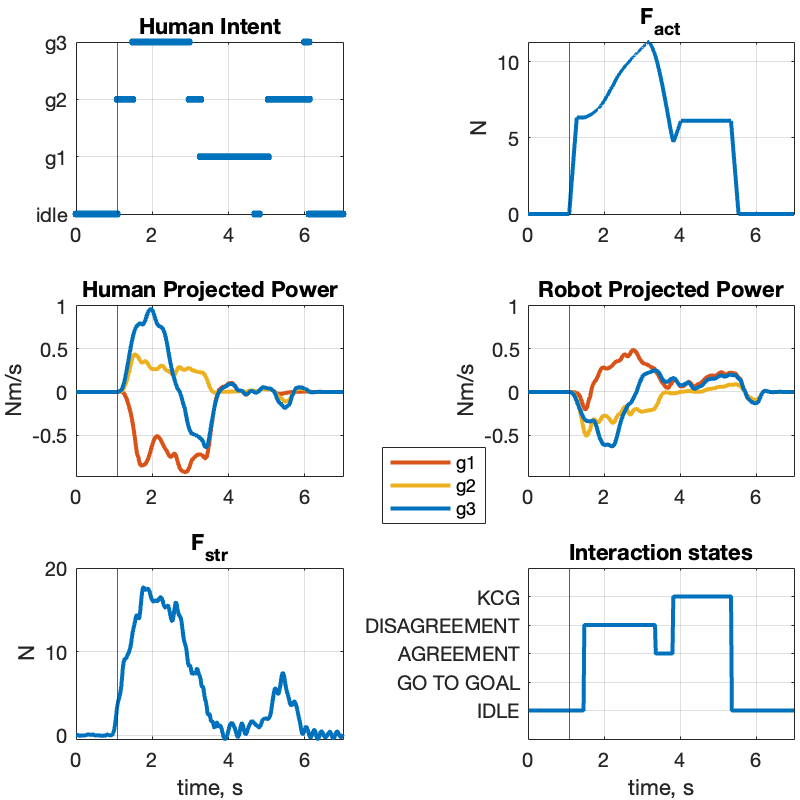}
    \caption{An example of soft-soft case where $g_R=g_1$ and $g_H=g_3$. The blue vertical curve shows the instant of the first beep signal.}
    \label{fig:soft_soft}
    \vspace{-5mm}
\end{figure}

An example run of the soft goal controller is given in Fig.~\ref{fig:soft_soft}. In this particular case, the intended goal for the human participant was $g_3$. In the human projected power curves, one can see clear intent towards $g_3$, which is correctly captured by the intent recognizer. Also, note the magnitude of $F_{str}$ that shows the level of the interaction conflict. In the conflicting phase, the controller correctly increases the $F_{act}$ in a disagreement mode and switches to the agreement phase as soon as the human partner decides to give up at the instant when the power curve drops sharply in magnitude (blue curve, 2nd row, left). Since the $F_{str}$ value did not cross the $F^C$ threshold, the robot did not give up on its intent. Finally, after a short agreement phase, the dyad follows the KCG mode and finishes the interaction. Video of the robot performance for all controllers is included in the supplemental material.

\subsubsection{Switching Frequency}


A valid concern in machine learning-based feedback controllers is fast switching due to misclassification. Therefore, we report the frequency of switches in the agreement-disagreement cycle and their average duration. In the hard goal controller (90 trials), the average duration spent in agreement and disagreement states were $1.37$s and $1.44$s, respectively. The average number of switches was $0.8$. In other words, the hard goal controller made one transition in 8 out of 10 cases, and no transitions in the remaining 2 cases. Similarly, in the soft goal controller (130 trials), the average duration of agreement and disagreement states were $1.6$s and $0.9$s, respectively. The average switching rate was $0.7$, which is less frequent compared to the hard goal controller. These results demonstrate the robustness of the proposed controller scheme against false transitions detected by the intent recognizer. Contributing factors to this are accurate tuning of timer thresholds (see Sec. \ref{sec:hlc}) and the precision of the intent recognizer.

\subsection{Survey Results}

In the follow-up survey, each participant was asked the following 5 questions:
\begin{enumerate*}
    \item How comfortable was the interaction? 
    \item How responsive was the robot?
    \item Was the robot’s behavior predictable?
    \item If you had to perform this task of moving an object with a partner over a longer period of time, would it be acceptable to have this robot as the partner?
    \item How similar was the robot’s behavior to human behavior?
\end{enumerate*}
For the response, we used a 7-point Likert scale, with 7 being the most positive response~\cite{schrumConcerningTrendsLikert2023}. The average values of responses for the 10 participants are summarized in Tab.~\ref{tab:likert_scale_average}. The results show that the participants very highly rated the interaction.

\begin{table}[h]
\centering
\resizebox{0.8\columnwidth}{!}{%
\begin{tabular}{lccccc}
\hline
\hline
Question \# & \textbf{Q1} & \textbf{Q2} & \textbf{Q3} & \textbf{Q4} & \textbf{Q5}\\ \hline
Average & 5.8 & 6.2 & 5.4 & 5.3 & 5.2 \\ \hline \hline
\end{tabular}
}
\caption{Average responses in Likert scale for 10 participants}
\label{tab:likert_scale_average}
\end{table}

\section{Conclusion}

The paper describes a hierarchical robot control framework that emulates human behavior in collaborative manipulation tasks. The framework is inspired by our study of human-human collaboration which showed how humans use force exchanges to reach a consensus on where to move, and who will take the lead role. At the top level, the proposed controller, inspired by these human strategies, consists of finite-state machines that represent different levels of commitment to a desired goal configuration. Key to our architecture is a real-time intent recognizer that enables the robot to respond to human actions effectively. We provide insights into the controller's design, discuss the feature engineering process for the intent recognizer, and describe the recognizer training.

The framework was implemented on a UR10e robot and evaluated through human studies. The experimental results demonstrate the robot's ability to correctly recognize and respond to human input, effectively communicate its intent, and resolve conflicts. Success rates are reported, and comparisons with human-human experiments are made to illustrate the approach's effectiveness in emulating human collaborative behavior in robot-human interactions. A qualitative evaluation of the controller by the human subjects shows that it was very favorably perceived.

A natural extension of this work is to adapt the controller in a realistic setting, where human and robot can co-manipulate various objects with different destination distributions. This will require an environment-invariant version of the presented intent recognizer. Another interesting path to explore is to expand the modality with the addition of language and gesture in the high-level controller.




\bibliographystyle{IEEEtran}
\bibliography{icra_2024,icra_mz}


\end{document}